\documentclass{article}

\usepackage{microtype}
\usepackage{graphicx}
\usepackage{subfigure}
\usepackage{booktabs} 
\usepackage{tikz}
\usepackage{pgfplots}
\usepackage{filecontents}
\usepackage{multirow}
\usepackage{rotating}
\usepackage{tabularx}

\usepackage{hyperref}

\usepackage[accepted]{icml2024}

\usepackage{amsmath}
\usepackage{amssymb}
\usepackage{mathtools}
\usepackage{amsthm}

\usepackage[capitalize,noabbrev]{cleveref}

\theoremstyle{plain}

\theoremstyle{definition}

\theoremstyle{remark}

\usepackage[textsize=tiny]{todonotes}

\begin{document}

\twocolumn[
\icmltitle{MIP: CLIP-based Image Reconstruction from PEFT Gradients
}



\icmlsetsymbol{equal}{*}

\begin{icmlauthorlist}
\icmlauthor{Peiheng Zhou}{ecnu}
\icmlauthor{Ming Hu}{ntu}
\icmlauthor{Xiaofei Xie}{smu}
\icmlauthor{Yihao Huang}{ntu}
\icmlauthor{Kangjie Chen}{ntu}
\icmlauthor{Mingsong Chen}{ecnu}
\end{icmlauthorlist}

\icmlaffiliation{ecnu}{East China Normal University, China}
\icmlaffiliation{ntu}{Nanyang Technological University, Singapore}
\icmlaffiliation{smu}{Singapore Management University, Singapore}

\icmlcorrespondingauthor{Mingsong Chen}{mschen@sei.ecnu.edu.cn}


\vskip 0.3in
]



\printAffiliationsAndNotice{}  

\begin{abstract}
Contrastive Language-Image Pre-training (CLIP) model, as an effective pre-trained multimodal neural network, has been widely used in distributed machine learning tasks, especially Federated Learning (FL).
Typically, CLIP-based FL adopts Parameter-Efficient Fine-Tuning (PEFT) for model training, which only fine-tunes adapter parameters or soft prompts rather than the full parameters.
Although PEFT is different from the traditional training mode, in this paper, we theoretically analyze that the gradients of adapters or soft prompts can still be used to perform image reconstruction attacks.
Based on our theoretical analysis, we propose Multm-In-Parvo (MIP), a proprietary reconstruction attack method targeting CLIP-based distributed machine learning architecture.
Specifically, MIP can reconstruct CLIP training images according to the gradients of soft prompts or an adapter.
In addition, MIP includes a label prediction strategy to accelerate convergence and an inverse gradient estimation mechanism to avoid the vanishing gradient problem on the text encoder.
Experimental results show that MIP can effectively reconstruct training images according to the gradients of soft prompts or adapters of CLIP models.

\end{abstract}

\section{Introduction}

Due to the support of text data to assist image recognition tasks, Contrastive Language-Image Pre-training (CLIP) models have shown advantages in various applications, such as Video Text Retrieval \cite{clip4clip, clipbert}, Artificial Intelligence Generated Content (AIGC) \cite{stable, dalle}, and generalized multimodal tasks \cite{audioclip, pointclip, depthclip}.
As a pre-trained model, the CLIP model is typically used as an initial model and then fine-tuned with personalized datasets for specific tasks.
Due to its strong generalization, CLIP models are widely used as an initial model in distributed machine learning frameworks~\cite{distribute1,distribute2,distribute3,distribute4}, especially Federated Learning (FL) \cite{FL}, to adapt to different personalized tasks.

Typically, in FL, the cloud server dispatches the global model to multiple clients for local training and each client uploads the gradients to the cloud server for aggregation.
In this way, clients can achieve collaborative model training without data sharing.
However, untrusted servers can still reconstruct training images according to uploaded gradients using Deep Leakage from Gradients (DLG) technology~\cite{DLG}.
Based on DLG, various studies~\cite{IDLG,GradInversion,LAMP} have been presented to reconstruct images using gradients.
Although secure aggregation~\cite{bonawitz2017practical} can be used to avoid DLG attacks, due to the large communication and time overhead, it is seriously limited.
Since DLG attacks focus only on image models and CLIP adopts a different structure, existing DLG methods cannot be used directly to attack CLIP-based FL.
\textit{To balance the effectiveness and security of CLIP-based FL systems, exploring whether the existing CLIP model is at risk of being attacked by DLG is an important issue in CLIP-based FL system design.}

\begin{figure}[h]
\vspace{-0.1in}
\centering
\includegraphics[width=0.47\textwidth]{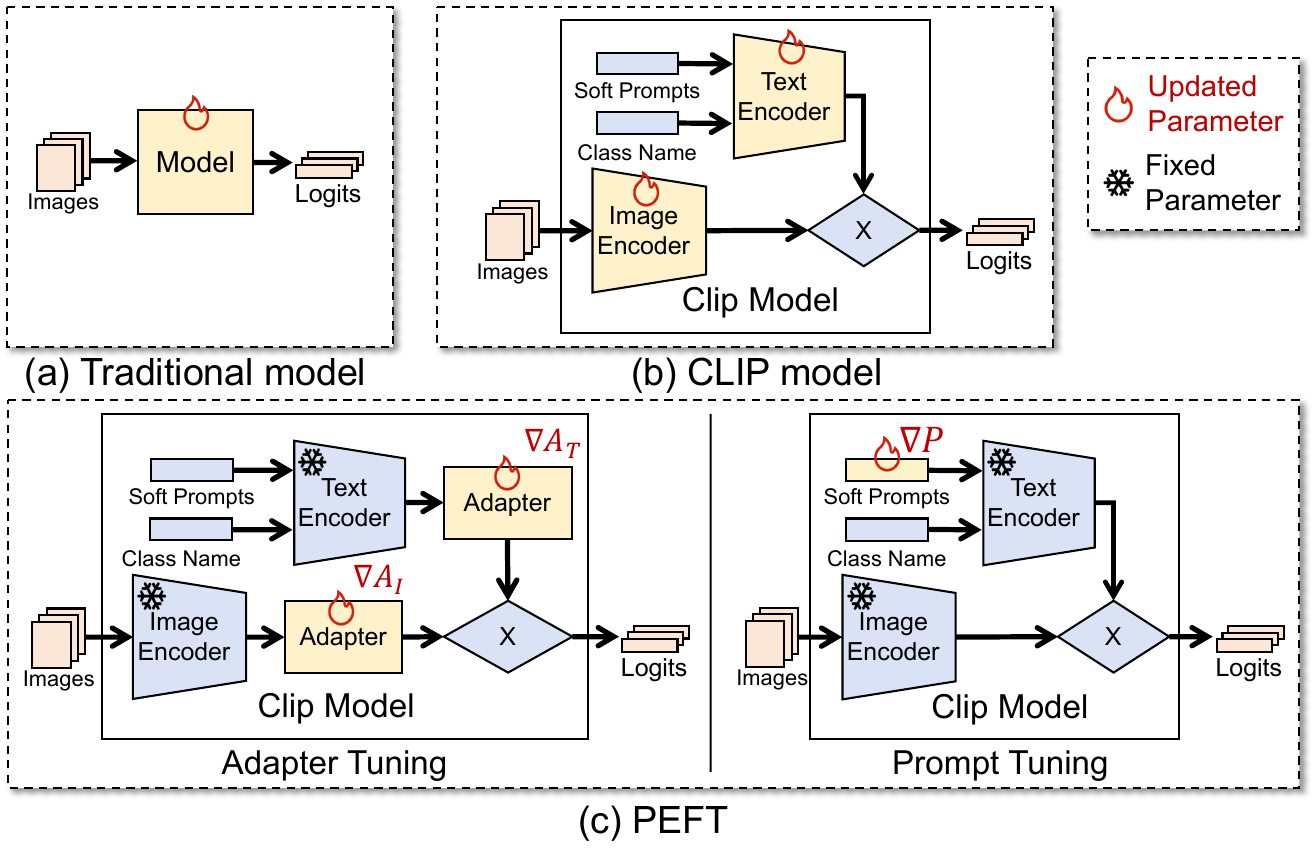}
\vspace{-0.25in}
\caption{Example of CLIP and PEFT.} 
\label{fig:intro_case}
\vspace{-0.2in}
\end{figure}

Figures~\ref{fig:intro_case} (a) and (b) present an example of a traditional image model and a CLIP model, respectively. Compared to the traditional model, CLIP has an additional text encoder and soft prompts.
When fine-tuning is performed, the parameters of both text and image encoders and soft prompts need to be updated.
Therefore, reconstructing the CLIP training images needs to take into account the gradients of the text encoder and soft prompts.
As the complexity of AI tasks increases, large CLIP models have gradually become a trend, resulting in unaffordable computation and communication overhead.
To achieve effective collaborative training of the CLIP model, Parameter-Efficient Fine-Tuning (PEFT)~\cite{PEFT} has become a trend in CLIP-based FL systems.
When PEFT is adopted, each client updates only a small portion of the parameters, and a large number of skeletal parameters are frozen.
Since CLIP is pre-trained, only training a small portion of the parameters can enable CLIP to learn a personalized task.
Figure~\ref{fig:intro_case} (c) presents an example of PEFT for the CLIP model.
For CLIP-based FL, each client trains only the text or image adapter or soft prompts and fixes all the other parameters.
Therefore, \textit{the goal of this paper is to reconstruct training images only according to gradients of adapters or soft prompts.}

In this paper, we represent the first attempt to apply image reconstruction attacks within the PEFT training mode. We propose Multm-In-Parvo(MIP), a reconstruction attack method based on CLIP-like Construction. We provide theoretical and experimental evidence to demonstrate the feasibility of reconstructing training data by extracting gradients from fine-tuned unfrozen parameters while disregarding other gradients, leading to privacy leakage. Furthermore, we develop a tailored reconstruction workflow that significantly enhances the quality of the recovered images and is better suited for research scenarios. Our main contributions are summarized as follows:
\begin{itemize}
\vspace{-0.1in}
\setlength\itemsep{-0.1em}
\item[$\bullet$] We transfer the idea of a reconstruction attack to the training mode based on efficient parameter tuning in multimodal scenes, and we demonstrate its feasibility through proof-of-concept. To our knowledge, it’s the first work of private data reconstruction for this application scenario.
\item[$\bullet$] We propose Multm-In-Parvo(MIP), which uses CLIP as the multi-modal backbone and designs a reconstruction attack method based on PEFT mode. Our method includes label prediction and image reconstruction.
\item[$\bullet$] We conduct extensive experiments and statistics to determine the success rate and quality of MIP's reconstruction on different categories of images. We also include ablation experiments to analyze the effects of each part of the method seriatim.
\end{itemize}

\section{Related Work}

\subsection{PEFT on CLIP}

CLIP \cite{CLIP}, a multimodal background LM that undergoes unsupervised pre-training through contrastive learning \cite{contrast}, projects semantic information and image feature information into the same space for computation. PEFT \cite{PEFT, adapter, prefix, ptuning, lora} is a kind of training methodology based on large models, typically freezing the majority of parameters within the model and fine-tuning a small subset of parameters through incremental training. Its integration with CLIP involves CoOp \cite{coop}, which introduces soft prompts to CLIP, replacing static templates and optimizes the tokens of prompts through few-shot incremental training, and CLIP-adapter \cite{clipadapter}, which incorporates adapters into CLIP, adding them as trainable modules after the feature encoders on both the image and text sides.

\subsection{Distributed Training and Privacy Protection}
Distributed training frameworks are typically divided into decentralized distribution \cite{decen1,decen2} or centralized distribution \cite{cen1,cen2}. The key distinction lies in the presence or absence of a centralized server that coordinates distributed computations and consolidates the computational results. Federated Learning (FL) \cite{FL} is a quintessential distributed training paradigm that maintains a cloud server, constituting a centralized mechanism. The integration of large models and FL has led to the emergence of PEFT in distributed training applications, such as FedPrompt \cite{FedPrompt} and FedAdapter \cite{FedAdapter}. In these types of architectures, updated information related to the PEFT module replaces gradients as the content is communicated with the server. On the other hand, FedClip \cite{FedClip} combines CLIP and adapter with FL, while pFedPrompt \cite{pfedprompt} designs an FL algorithm for transferring prompts based on the CoOp mode.

\subsection{Reconstruction Attack}
The pattern of exchanging gradients only or a small subset of critical information between training nodes in FL was previously considered secure. However, DLG \cite{DLG} proposed a strategy based on gradient reconstruction of training data and demonstrated its feasibility. IDLG \cite{IDLG} addressed the issues of poor convergence quality and suboptimal reconstruction effects under a high number of batches by proposing label prediction. Inverting Gradients \cite{InvertingGradients} modified the optimization loss and measured reconstruction efficiency under different training environments. GradInversion \cite{GradInversion} attempted to alleviate the negative impact of high batch numbers through additional regularization terms and combination strategies. LAMP \cite{LAMP} introduced natural language semantic optimization based on GPT-2 and formed alternate optimization with the original process aimed at text reconstruction. TAG \cite{TAG} also focuses on this direction but extends the attacked model to transformers. GGL \cite{GGL} used prior knowledge to assist in the optimization direction of dummy data in the reconstruction field, leading the way in terms of superiority.

The aforementioned methods primarily concentrate on scenarios involving unimodal and small models, with some also modestly incorporating prior knowledge from large models. However, there is an absence of research precedents for reconstruction attacks under the multimodal and PEFT training framework, resulting in a lack of theoretical guidance for the implementation of reconstruction tasks. Our approach has accomplished scenario transfer, theoretically substantiating its feasibility, and further proposes several optimization steps to adapt to the scenario.

\section{Inverting Gradients from PEFT}

\subsection{Condition and Environment of Attack}
The environment upon which our work is based conforms to commonly used standards within the field. Reconstruction attacks are implemented on a malicious server or scheduling center, with the source of the malicious attack being driven by curiosity rather than intent to tamper with information. This implies that the malicious attacker is able to invoke a general model under a centralized framework and obtain information about its parameters but is unable to access any private client data. By intercepting communication contents, the attacker can obtain local update gradients transmitted from clients to the centralized server, which represent the only content that an attacker can acquire from clients. In PEFT training mode, most model parameters are frozen, with only a few parameters left for fine-tuning and generating gradients. Therefore, the transmission content from clients concerns gradients about those small and hot network modules. Additionally, extra computational overhead and open-source prior knowledge can be leveraged.

In contrast to conventional reconstruction attacks, MIP is conducted on a CLIP-like construction, which means that the attacked model is used for a multimodal classification task involving text-image inference. The CLIP architecture extracts features from two modalities of data by setting up text and image encoders, which can be arbitrarily replaced regardless of whether they are pre-trained or not. For example, encoder pairs such as transformer\&Resnet or transformer\&vision-transformer can be utilized. For text data, the input encoder includes a set of prompts as well as natural language names corresponding to the labels in the dataset. After encoding, both modalities' data are normalized and aligned before being multiplied together to generate the inference result. Therefore, the core objective of MIP is to reconstruct the training images by obtaining a small portion of gradients primarily originating from the text feature module.

\subsection{Transfer of Optimization Problem}
\label{transfer}
After the major architecture changes in the model, the optimization principle of the reconstruction attack needs to be re-derived. The purpose of the reconstruction attack is to optimize the equation as follows:
{
\setlength\abovedisplayskip{0.2cm}
\setlength\belowdisplayskip{0.1cm}
\begin{eqnarray}
\label{eq1}
X = \underset{X}{argmin} \ \mathcal{L}(X;\nabla \mathcal{W},\nabla \mathcal{W}^{\prime}).
\end{eqnarray}
}
When it comes to CLIP-like construction, the optimization purpose will convert to: 
{
\setlength\abovedisplayskip{0.2cm}
\setlength\belowdisplayskip{0.1cm}
\begin{eqnarray}
\label{eq2}
X = \underset{X}{argmin} \ \mathcal{L}(X;\nabla\mathcal{P},\nabla\mathcal{P}^{\prime}) \ \ s.t. \ 
 \mathcal{P} = \mathcal{W} - \mathcal{F}.
\end{eqnarray}
}
The F means frozen parameters from the model, while P means hot parameters from the set of W.

In order to adhere to the prior theoretical framework, we performed iterative optimization using the expression method presented as follows: 
{
\setlength\abovedisplayskip{0cm}
\setlength\belowdisplayskip{0cm}
\begin{eqnarray}
\label{eq3}
X_{i+1} = X_{i} - \eta\nabla_{X_{i}} \mathcal{L} \ \ , \ \nabla_{X_{i}} \mathcal{L} =\frac{\partial \mathcal{L}(\nabla \mathcal{P},\nabla\mathcal{P}^{\prime})}{\partial X_{i}}.
\end{eqnarray}
}

Prior to optimization, a noise X was initialized, whose pixels could conform to any random distribution and had the same dimensions as the target image to be restored, which could be obtained from the input dimension parameters from the cloud server. The updated content for each iteration consists of the learning rate $\eta$ multiplied by the gradient of the loss $\mathcal{L}$ with respect to the dummy image X.

In Eq.~\ref{eq3}, $\nabla\mathcal{P}$ is obtained from communication with the client and is an irrelevant quantity. $\nabla\mathcal{P}^{\prime}$ is the update obtained after training the model using the dummy X. The objective of the reconstruction attack is to make $\nabla\mathcal{P}^{\prime}$ as close as possible to $\nabla\mathcal{P}$. By structuring $\nabla\mathcal{P}^{\prime}$ based on Logits divisions, we can derive the following expanding:
{
\setlength\abovedisplayskip{0.2cm}
\setlength\belowdisplayskip{0cm}
\begin{eqnarray}
\label{eq4}
\nabla\mathcal{P}^{\prime} = \frac{\partial L(Y,GT)}{\partial\mathcal{P}^{\prime}} = \frac{\partial L(Y,GT)}{\partial Y}  \frac{\partial Y}{\partial \mathcal{P}^{\prime}}.
\end{eqnarray}
}

In CLIP, the definition of Y, the logits of the model is the product of multimodal feature alignment, which includes IF (Image Feature), TF (Text Feature), and a coefficient constant LS (Logit Scale), which can be expressed as:
{
\setlength\abovedisplayskip{0.2cm}
\setlength\belowdisplayskip{0.1cm}
\begin{eqnarray}
\label{eq5}
Y = LogitScale * ImageFeature * TextFeature^{T}.
\end{eqnarray}
}
Substituting the aforementioned expansion into the optimization gradient yields the following equation:
{
\setlength\abovedisplayskip{0.2cm}
\setlength\belowdisplayskip{0.1cm}
\begin{eqnarray}
\label{eq6}
\nabla_{X} \mathcal{L} = [\frac{\partial L(Y,GT)}{\partial Y}\frac{\partial (LS*IF*TF^{T})}{\partial \mathcal{P}^{\prime}}]\bigg|_{X}^{\prime},
\end{eqnarray}
}
Which is obviously different from the optimization objectives of traditional scenarios:
{
\setlength\abovedisplayskip{0.2cm}
\setlength\belowdisplayskip{0.1cm}
\begin{eqnarray}
\label{eq7}
\nabla_{X} \mathcal{L} = [\frac{\partial L(Y,GT)}{\partial Y}\frac{\partial F(X;\mathcal{W})}{\partial \mathcal{W}^{\prime}}]\bigg|_{X}^{\prime}.
\end{eqnarray}
}

\emph{\textbf{Proposition 3.2:}\label{prop1} Consider a classification NN that utilizes  PEFT  and incorporates bidirectional feature alignment multiplication. A reconstruction attack targeting the leakage of gradients in modules can be reduced to the DLG problem with identical theoretical convergence properties. Building upon this, image reconstruction in the new scenario becomes a first-derivative problem, thereby widening the scope of attacks and exhibiting a layer-free nature.}

To break down Eq.~\ref{eq6} and Eq.~\ref{eq7}, both terms are subject to two rounds of differentiation. In comparison to the optimization gradient for stealing single-modal parameters (Eq.~\ref{eq7}), in the transferred approach, when the network generates updates, it optimizes with respect to the prompt by taking the gradients along the text pathway, as shown in this equation:
{
\setlength\abovedisplayskip{0.1cm}
\setlength\belowdisplayskip{0.1cm}
\begin{eqnarray}
\label{eq8}
\nabla_{\mathcal{P}^{\prime}} Y = LS*IF*\frac{\partial TF^{T}}{\partial\mathcal{P}^{\prime}}.
\end{eqnarray}
}
During reconstruction, optimization is performed with respect to the dummy X by taking the gradients along the image pathway. Clearly, the gradient paths for the two rounds of differentiation are not reused, as follows:
{
\setlength\abovedisplayskip{0.1cm}
\setlength\belowdisplayskip{0.1cm}
\begin{eqnarray}
\label{eq9}
\nabla_{\mathcal{P}^{\prime}} Y \big|_{X}^{\prime} = LS*\frac{\partial IF}{\partial X}*\frac{\partial TF^{T}}{\partial\mathcal{P}^{\prime}}.
\end{eqnarray}
}
According to the known conclusion that DLG is a second-derivative problem \cite{DLG}, which implies that in order to achieve successful attacks, the target NN should not have any layers that are non-differentiable at the second-derivative. Therefore, in practical experiments, layers such as ReLU are replaced. However, in the multi-modal scenario of this paper, this constraint no longer exists. For NN using any conventional structure, the effectiveness of the attack only depends on the depth of the NN and the overall scale of gradient variations it generates, making it layer-free. For the specific case of PEFT, when using soft prompts, the formulation is given by: 
{
\setlength\abovedisplayskip{0cm}
\setlength\belowdisplayskip{0.1cm}
\begin{eqnarray}
\label{eq12}
\nabla_{P^{\prime}} Y \big|_{X}^{\prime} = LS*\frac{\partial IF}{\partial X}*\frac{\partial TF^{T}}{\partial TE}*\frac{\partial TE}{\partial P^{\prime}},
\end{eqnarray}
}
When using an adapter, the formulation is given by:
{
\setlength\abovedisplayskip{0.1cm}
\setlength\belowdisplayskip{0.1cm}
\begin{eqnarray}
\label{eq10}
\nabla_{A^{\prime}} Y \big|_{X}^{\prime} = LS*\frac{\partial IF}{\partial X}*\frac{\partial TF^{T}}{\partial A^{\prime}}.
\end{eqnarray}
}
It should be noted that when maintaining the gradient along the Text Feature pathway, the adapter can only be used after the text encoder. Conversely, both the text and image encoders are used before the image adapter, as follows:
{
\setlength\abovedisplayskip{0.1cm}
\setlength\belowdisplayskip{0.1cm}
\begin{eqnarray}
\label{eq11}
\nabla_{A^{\prime}} Y \big|_{X}^{\prime} = LS*(\frac{\partial^{2} IF}{\partial A_{I}^{\prime}\partial X}*TF^{T} + \frac{\partial IF}{\partial X}*\frac{\partial TF^{T}}{\partial A_{T}^{\prime}}),
\end{eqnarray}
}
In this case, the image encoder can remain layer-free while the adapter is under second-derivative, making the original conclusion not tenable inside.

\section{MIP: Our Methodology}
\vspace{-3pt}
\subsection{Motivation}
\vspace{-3pt}
\label{motiva}
The application of the multimodal large-scale model fine-tuning method in a distributed framework brings changes to the communication gradient and training mode in the original scenario. Therefore, we found that if we only perform a simple migration of DLG without incorporating other content, i.e., optimizing Eq.~\ref{eq13}, it is likely that no reconstruction results will be obtained in practical scenarios.

The most crucial impact is that when maintaining the first-order optimization problem outside of Eq.~\ref{eq11}, where gradients come from the textual modality, the derivative chain of the optimization method becomes longer, significantly affecting the quality of reconstruction and even greatly reducing the convergence probability. Additionally, when involving the backward computation of large-scale encoders deployed on two modalities, phenomena such as gradient vanishing seem difficult to avoid, leading to a success rate of 0 for raw transferred DLG attacks that rely on stealing soft prompts. Therefore, we propose the MIP method to address these new problems arising in new scenarios. As the first reconstruction attack strategy applied in the multimodal domain, we also need to quantitatively test the impact of more complex gradient chains on restored image quality.

\begin{figure}[t]
\centering
\includegraphics[width=0.49\textwidth]{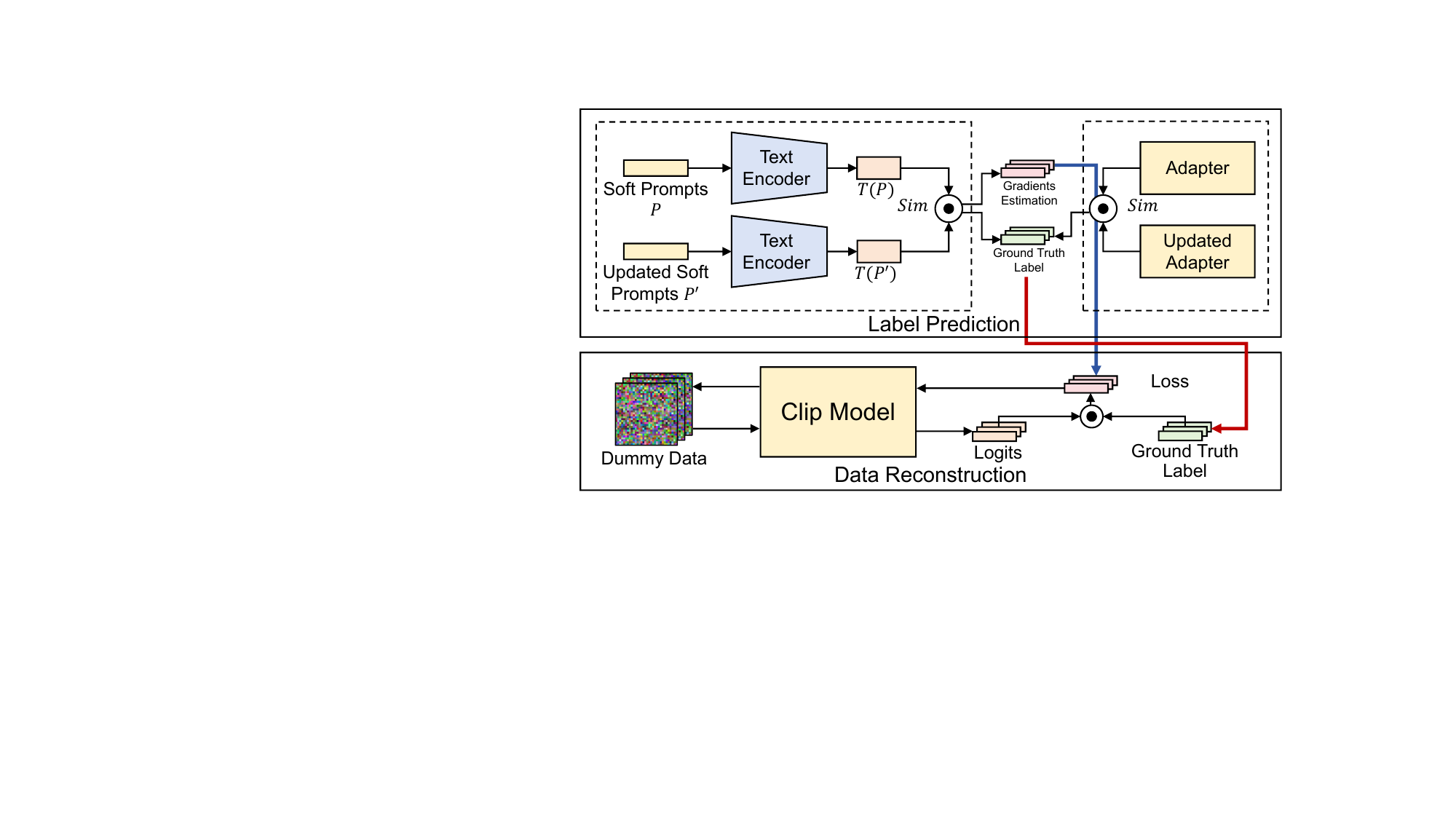}
\vspace{-0.25in}
\caption{Workflow of MIP} 
\label{flowchart}
\vspace{-0.15in}
\end{figure}

\subsection{Overview of MIP}

Figure~\ref{flowchart} shows an overview of MIP that utilizes model parameters visible within the server and some prior knowledge to construct a specialized reconstruction attack process for the CLIP framework, including label prediction and image reconstruction. Specifically, MIP consists of two main components: first, it guesses the ground truth corresponding to the training image based on the stolen PEFT gradients and simplifies the optimization target to the dummy image. Then, different strategies for image reconstruction are employed based on the PEFT pattern. For the possible vanishing gradient problem, MIP adopted specific strategies. These two components are discussed in detail below.

\subsection{Label Prediction with gradients of PEFT}

Label prediction is an important strategy for improving the convergence efficiency of reconstruction. Without deploying it, the optimization objective would degenerate into:
{
\setlength\abovedisplayskip{0.2cm}
\setlength\belowdisplayskip{0.1cm}
\begin{eqnarray}
\label{eq13}
X,Y = \underset{X,Y}{argmin} \ \mathcal{L}(X,Y;\nabla\mathcal{P},\nabla\mathcal{P}^{\prime}).
\end{eqnarray}
}
Utilization of Eq.~\ref{eq13} leads to increased difficulty in finding the convergence interval within the search space in our multimodal structure. IDLG proposed label guessing for the first time, and it has a transferable conclusion \cite{IDLG} as follows:
{
\setlength\abovedisplayskip{0.1cm}
\setlength\belowdisplayskip{0.2cm}
\begin{eqnarray}
\label{eq14}
\frac{\partial L(Y,GT)}{\partial y_{i}} = \left\{\begin{matrix} 
  -1 + \frac{e^{y_{i}}}{\Sigma_{j}e^{y_{j}}}\ , \ if \  i=GT \\  \\
  \frac{e^{y_{i}}}{\Sigma_{j}e^{y_{j}}} \ , \ else
\end{matrix}\right. .
\end{eqnarray}
}
Specifically, when using cross-entropy as a loss function, the gradient of the component corresponding to the ground truth is negative, while the gradients of other components are positive. The approach employed by IDLG for label prediction involves comparing and judging based on the gradients obtained from the fully connected layers of the network, as shown in the following equation:
{
\setlength\abovedisplayskip{0.2cm}
\setlength\belowdisplayskip{0.2cm}
\begin{eqnarray}
\label{eq15}
GT = i \ , \ s.t. \ \nabla {\mathcal{W}_{L}^i}^T * \nabla \mathcal{W}_{L}^j \leq 0 \ , \ \forall j \neq i \ .
\end{eqnarray}
}
However, in the context of MIP, it is not possible to obtain gradients from the final layers of the network, and the original form of judgment is also inconsistent with the CLIP framework. Therefore, it is necessary to revalidate and provide new proof for the revised approach.

MIP uses the gradient estimate of the text feature, denoted as $\nabla_{TF}L(Y,GT)$, to replace the end gradient $\nabla\mathcal{W}_L$ for executing ground truth judgment because when PEFT module is on the text-side path, the gradient estimate of TF can be obtained through $\nabla\mathcal{P}$, while the information from IF-side is unknown. To solve for $\nabla_{TF}L(Y,GT)$, the update of TF, denoted as TF', is obtained through the update of $\nabla\mathcal{P}$. When it is soft prompts, the estimate can be expressed as:
{
\setlength\abovedisplayskip{0.1cm}
\setlength\belowdisplayskip{0.1cm}
\begin{eqnarray}
\label{eq16}
TF^{\prime} = TE(P-\eta\nabla P),
\end{eqnarray}
}
When it is a text adapter, the estimate will convert to:
{
\setlength\abovedisplayskip{0.1cm}
\setlength\belowdisplayskip{0.1cm}
\begin{eqnarray}
\label{eq17}
TF^{\prime} = A^{\prime}(TE(TextInput);A-\eta\nabla A).
\end{eqnarray}
}
We can provide the reverse calculation for estimating $\nabla_{TF}L$, and under the ordinary gradient optimization method without additional information, this estimation is accurate, which can be expressed as follows:
{
\setlength\abovedisplayskip{0.1cm}
\setlength\belowdisplayskip{0.1cm}
\begin{eqnarray}
\label{eq18}
\nabla_{TF}L(Y,GT) = \frac{1}{\eta}(TF - TF^{\prime}).
\end{eqnarray}
}
The different components in $\nabla_{TF}L$ can be combined to form an expression as Eq.~\ref{eq19} and simplify into Eq.~\ref{eq20}:
{
\setlength\abovedisplayskip{0.1cm}
\setlength\belowdisplayskip{0cm}
\begin{eqnarray}
\label{eq19}
\nabla_{TF_i}L^T * \nabla_{TF_j}L = {(\frac{\partial L}{\partial y_i}*\frac{\partial y_i}{\partial TF_i})}^T(\frac{\partial L}{\partial y_j}*\frac{\partial y_j}{\partial TF_j})
\end{eqnarray}
}
{
\setlength\abovedisplayskip{0cm}
\setlength\belowdisplayskip{0.1cm}
\begin{eqnarray}
\label{eq20}
= \frac{\partial L}{\partial y_i}^T*\frac{\partial L}{\partial y_j}*LS^2*IF^T*IF.
\end{eqnarray}
}
Obviously, in the various terms of the equation, $LS^2>0$, $IF^T*IF>0$. When the GT component is multiplied with a non-GT component, $\nabla_{TF_i}L^T * \nabla_{TF_j}L<0$, conversely, $\nabla_{TF_i}L^T * \nabla_{TF_j}L>0$, this can be mathematically expressed as follows:
{
\setlength\abovedisplayskip{0.1cm}
\setlength\belowdisplayskip{0.1cm}
\begin{eqnarray}
\label{eq21}
GT = i \ , \ s.t. \ \nabla_{TF_i}L^T * \nabla_{TF_j}L < 0 \ , \ \forall j \neq i.
\end{eqnarray}
}
The transferability of IDLG has been demonstrated in the aforementioned context. MIP employs a reverse gradient estimation to substitute the parameters at the end of the network. When predicting the label for the target image, given that the optimization rules align with the aforementioned assumptions, the prediction of the ground truth is strictly accurate. Through this process, the dummy label is transformed into a constant, simplifying the optimization objective to only the dummy image.

\subsection{Method of Reconstruction Attack}
\label{methodz}

In the image reconstruction module, it is necessary to consider the specific implementation of PEFT. MIP has implemented attacks on two types of PEFT structures: soft prompt before the text encoder \cite{coop} and text adapter after double encoders \cite{clipadapter}.

The Feasibility has been demonstrated in Section~\ref{transfer}, and it has been established that MIP conforms to the optimization objective resembling Eq.~\ref{eq2}. However, during the reconstruction process, a significant impact on the quality of reconstruction has been observed due to the phenomenon of gradient vanishing generated by the large model architecture. This is primarily manifested in the convergence of statistical probability and the degree of recovery of the dummy image. Based on experimental summaries, we propose the following assessment of the impact of the large model architecture on reconstruction attacks across various modalities under the CLIP framework:

\emph{\textbf{Proposition 4.4:}\label{prop2} The issue of gradient vanishing along the text modality path can be mitigated through reverse estimation, thus rendering the scale of the text encoder unrelated to the quality of reconstruction. However, the same problem cannot be circumvented along the data modality path. As the complexity of the image encoder increases, the quality of reconstruction tends to deteriorate.}

\textbf{Vanishing Gradient on Text Encoder.}
When using a text adapter, the gradient path for this module does not pass through the text encoder; hence, the issue of gradient vanishing does not exist. The problem discussed in this section is only targeted in the case of soft prompts. According to Eq.~\ref{eq16} and Eq.~\ref{eq18}, a combination formula is obtained:
{
\setlength\abovedisplayskip{0.1cm}
\setlength\belowdisplayskip{0.1cm}
\begin{eqnarray}
\label{eq22}
\nabla_{TF}L(Y,GT) = \frac{1}{\eta}(TF - TE(P-\eta\nabla P)).
\end{eqnarray}
}
Meanwhile, the gradient generated by the dummy image on the text feature in the cloud can be expressed as follows:
{
\setlength\abovedisplayskip{0.1cm}
\setlength\belowdisplayskip{0.1cm}
\begin{eqnarray}
\label{eq23}
\nabla_{TF^{\prime}}L(Y^{\prime},GT) = \frac{\partial L}{\partial Y^{\prime}}*LS*IF,
\end{eqnarray}
}
where $Y$ denotes the logits of the client, and $Y^{\prime}$ denotes the logits of the dummy image. Eq.~\ref{eq22} computes a constant gradient as the optimization target, while Eq.~\ref{eq23} computes the gradient to be optimized, thus circumventing the computation involving the text encoder.

After obtaining the above information, it is possible to reformulate the optimization objective as: 
{
\setlength\abovedisplayskip{0.1cm}
\setlength\belowdisplayskip{0.1cm}
\begin{eqnarray}
\label{eq24}
X_{i+1} = X_{i} - \eta\nabla_{X_{i}} \mathcal{L} \  ,  \nabla_{X_{i}} \mathcal{L} =\frac{\partial \mathcal{L}(\nabla TF,\nabla {TF^{\prime}})}{\partial X_{i}}.
\end{eqnarray}
}
This reveals the conclusion that there is no need to perform reverse operations on the large model along the gradient path. For the text feature, there are no other operations apart from normalization, which is theoretically an equivalent position of gradient.

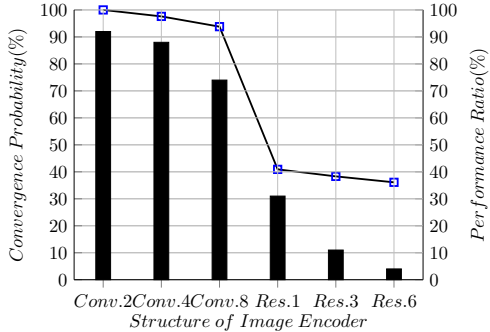
\begin{figure}[t]
\centering
\begin{tikzpicture}[scale=0.7]
\begin{axis}[
    xlabel=$x$,
    ylabel=$Performance~Ratio(\%)$,
    ylabel style={font=\small},
    ylabel near ticks,
    ymin=0,
    ymax=100,
    xmin=0.5,
    xmax=6.5,
    axis y line*=right,
    hide x axis,
    width=6cm,
    height=5cm,
    grid=major,
    clip=false,
    xtick={1,2,...,6},
    ytick={0,10,...,100},
    legend style={
      legend pos=north east,
      yshift=-20pt,
    },
    legend cell align=left,
    restrict y to domain=0:101,
    restrict x to domain=0:7,
    scale=1.5,
    axis on top=false,
]
\addplot [mark=square, mark options={blue}, line width=1pt] table [x=Index, y=Value2, col sep=comma] {data.csv};
\end{axis}
\begin{axis}[
    xlabel=$Structure~of~Image~Encoder$,
    ylabel=$Convergence~Probability(\%)$,
    ylabel style={font=\small},
    ylabel near ticks,
    ymin=0,
    ymax=100,
    xmin=0.5,
    xmax=6.5,
    axis lines*=left,
    width=6cm,
    height=5cm,
    grid=major,
    clip=false,
    xtick={1,2,...,6},
    xticklabels={$Conv.2$,$Conv.4$,$Conv.8$,$Res.1$,$Res.3$,$Res.6$},
    ytick={0,10,...,100},
    legend style={
      legend pos=north east,
      yshift=0pt,
    },
    legend cell align=left,
    restrict y to domain=0:100,
    restrict x to domain=0:7,
    scale=1.5,
    ybar,
    bar width = 8pt
]
\addplot [fill=black] table [x=Index, y=Value1, col sep=comma] {data.csv};
\end{axis}
\end{tikzpicture}

\vspace{-5pt}
\caption{Impact of encoder structure on reconstruction quality.}
\vspace{-10pt}
\label{stat}
\end{figure}

\textbf{Vanishing Gradient on Image Encoder.}
Unlike the text feature path, gradients through the image encoder cannot be replaced by reverse estimation because even if an estimation of the image feature is obtained, it still requires computation through the encoder. The optimization of the dummy image based on image feature updates cannot avoid the problem of gradient vanishing.

The phenomenon of gradient vanishing along the image path follows common patterns and is related to the scale of the encoder and the characteristics of its individual layers in intercepting gradients without the probability of optimizing. To address this, as shown in Figure~\ref{stat}, we conducted statistical experiments to estimate the impact of the encoder's structure on recovery quality and convergence probability.

In Figure~\ref{stat}, bars represent convergence probability, while polylines indicate the average reconstruction quality based on PSNR, with the value for the smallest structure set as 100\%. We selected six different NN structures, including three kinds of convolutional layers plus activation layers and three kinds of residual blocks, identical to CLIP. We computed the average results over 100 randomly selected images on each structure. 

Statistical results indicate that within neural networks composed of convolutional layers, the probability of convergence and image quality exhibit a quasi-linear decline as the depth of layers increases. Although this phenomenon can be mitigated by incorporating the loss from an image adapter, attacks against large-scale image encoders remain challenging to succeed. The residual block, with its more complex structure due to the inclusion of mechanisms such as attention, experiences a significant drop in performance on the same parameter scale. Even with a minimal number of residual blocks, it is almost impossible to reconstruct high-quality images, presenting an unavoidable challenge.

\section{Experiments}
We implemented our method, i.e., MIP, to evaluate its effectiveness. Our experimental content is based on reconstruction attacks utilizing a single image, with images from multiple datasets serving as the targets for restoration. Furthermore, we also incorporated ablation studies to ascertain the independent effectiveness of each module within MIP as well as the validity of transfer conclusions.

\subsection{Experimental Setup}

Our experiments are conducted in a PyTorch environment on an Ubuntu system. All experiments are completed on an Intel i7-13700K CPU and a single RTX 4090 GPU. 

\textbf{Dataset and Model Settings.} The experiments utilized data selected from Caltech101, CIFAR10, and MNIST as simulated training data for reconstruction. All images are resized to a fixed size of 28x28, 32x32, or 64x64 pixels, depending on their datasets. We conducted our method based on the CLIP architecture, with several different structures of encoders. In addition to the pre-trained module provided by CLIP, our statistical experiments also incorporate shallow perceptrons and a reduced number of residual blocks.

\textbf{Metrics.} We employ two metrics, Peak Signal-to-Noise Ratio (PSNR) and Structural Similarity Index (SSIM), as evaluation criteria for restored image quality. The calculation of PSNR is based on the definition of Mean Squared Error (MSE), measuring the error between the restored image and the original image at the pixel level. On the other hand, SSIM is predicated on the assumption that the human eye extracts structured information from an image, making it more widely utilized in the evaluation of image quality.

\begin{figure}[h]
\centering
\includegraphics[width=0.48\textwidth]{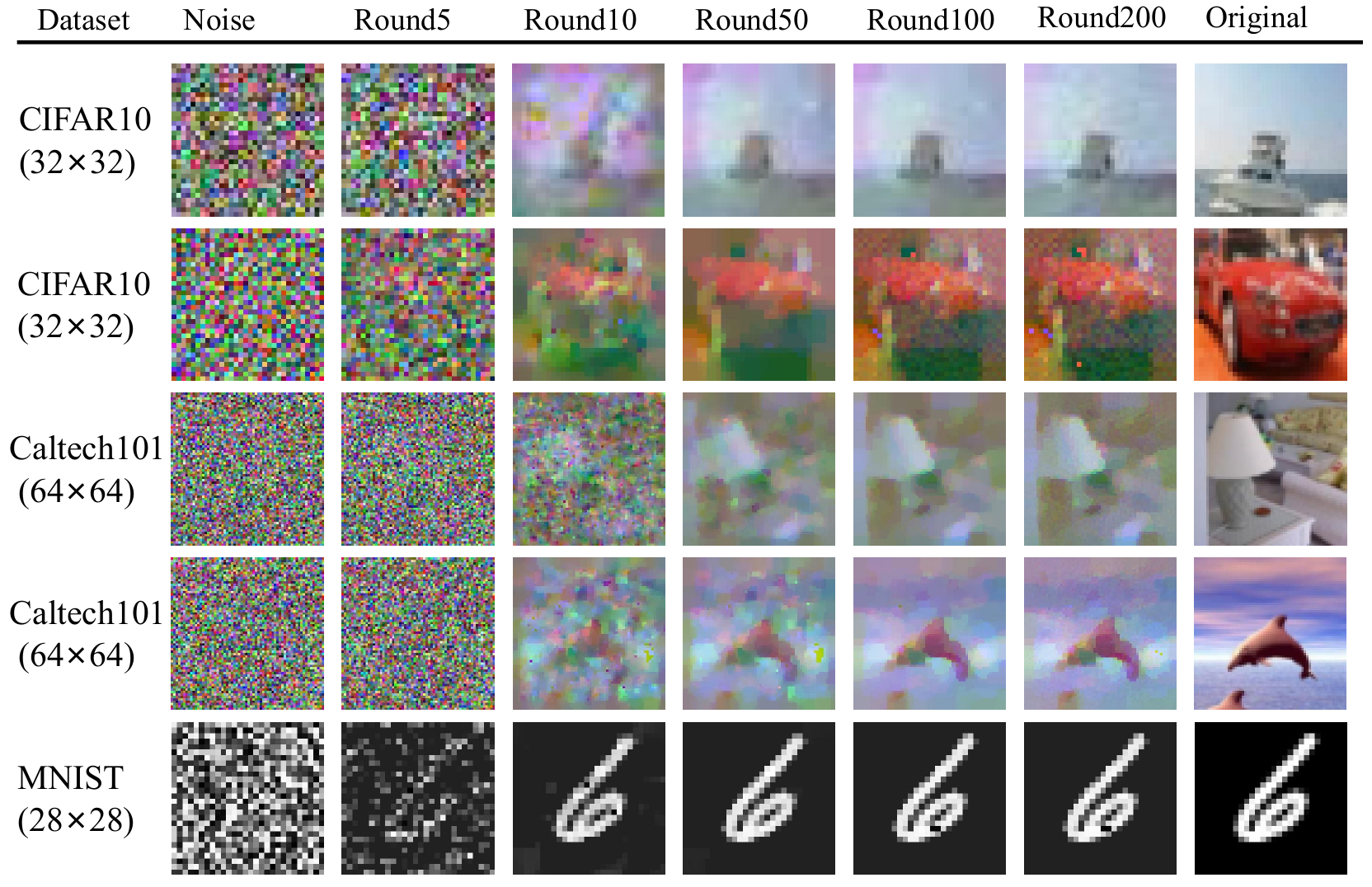}
\vspace{-20pt}
\caption{Reconstruction results based on attacks on soft prompts.} 
\label{ex1}
\end{figure}

\begin{figure}[h]
\centering
\includegraphics[width=0.48\textwidth]{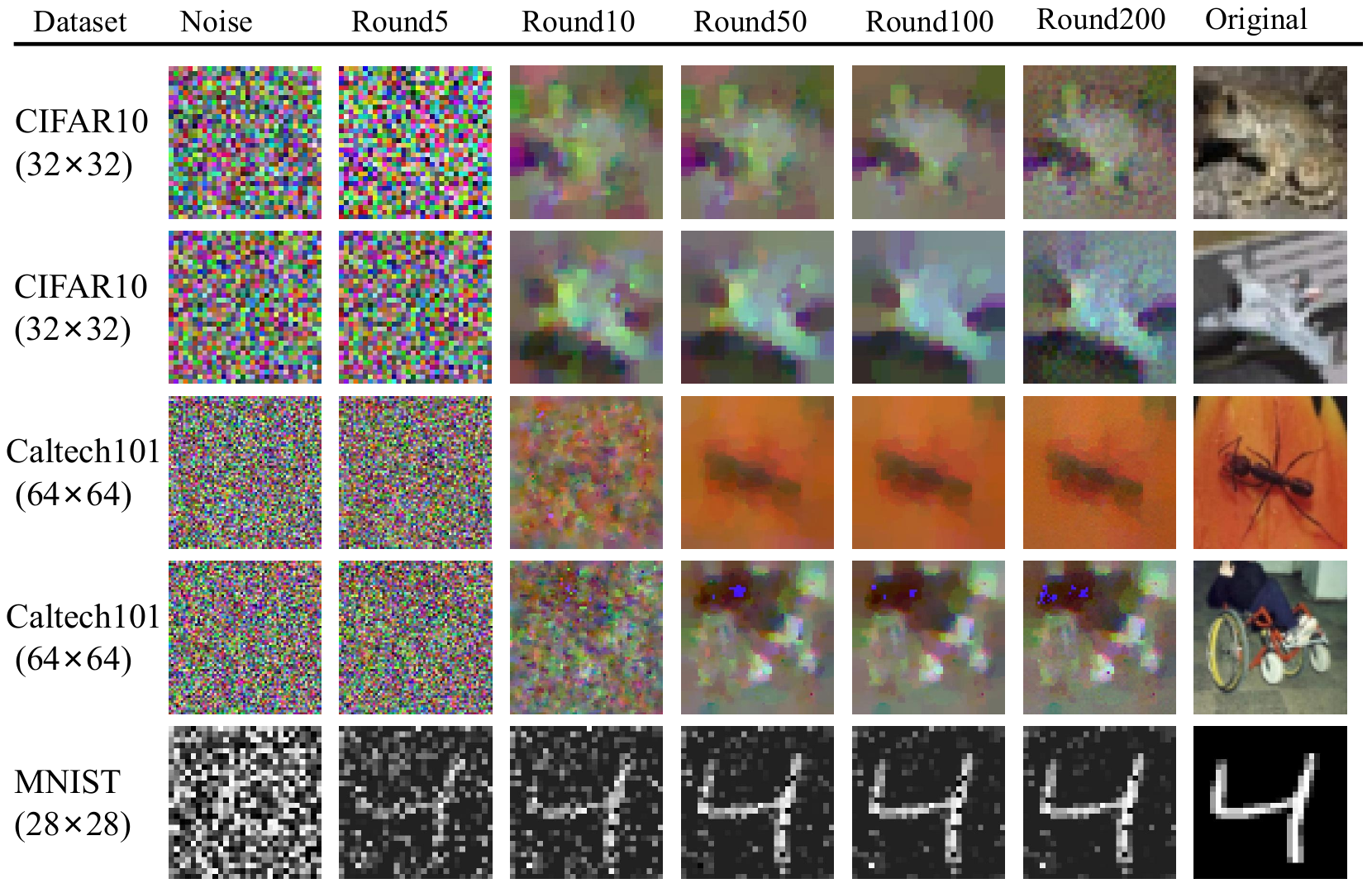}
\vspace{-20pt}
\caption{Reconstruction results based on attacks on text adapter.} 
\label{ex2}
\vspace{-5pt}
\end{figure}

\begin{figure}[h]
\centering
\includegraphics[width=0.48\textwidth]{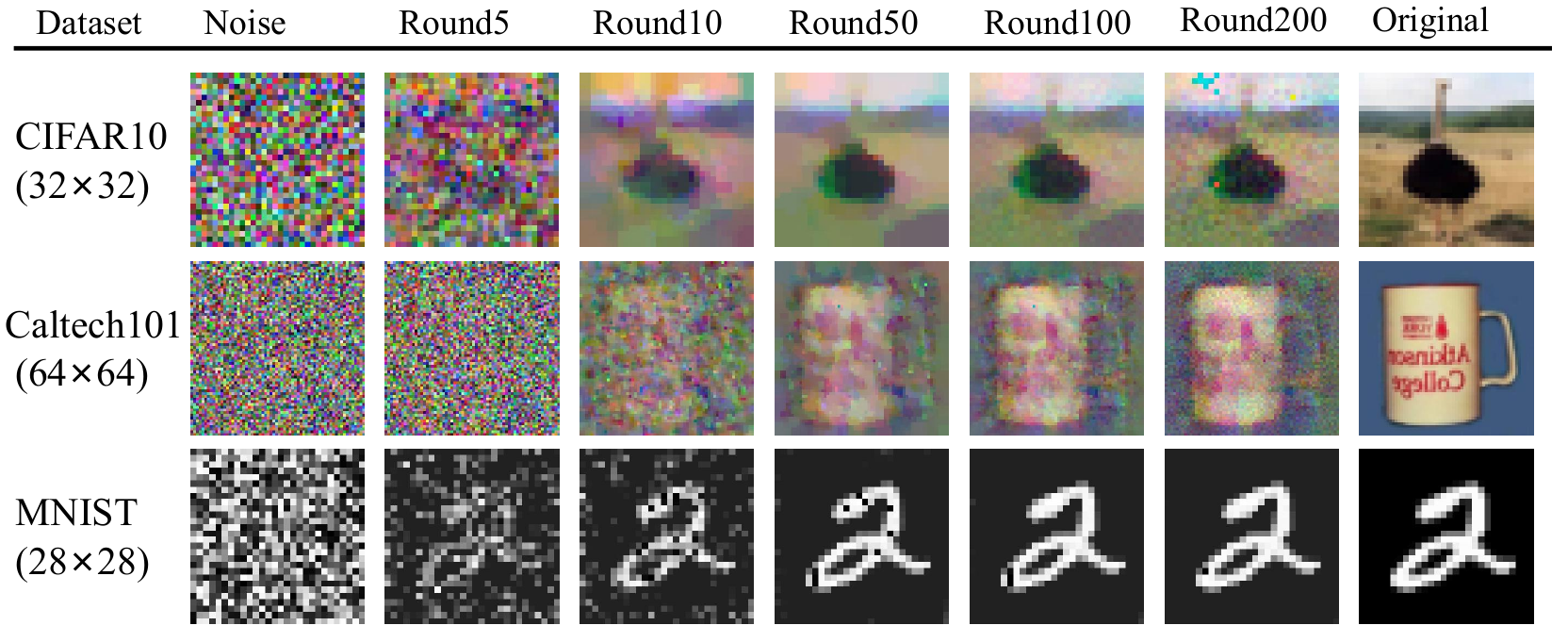}
\vspace{-20pt}
\caption{Reconstruction results of attacks on double adapters.} 
\label{ex3}
\vspace{-10pt}
\end{figure}

\subsection{Experiment Results}

The overall optimization objective of the experiment was derived from Section 3.2. To accelerate convergence and improve reconstruction quality during the actual iterative process, TV-Loss was added as follows:
{
\setlength\abovedisplayskip{0.1cm}
\setlength\belowdisplayskip{0cm}
\begin{eqnarray}
\label{eq25}
X = \underset{X}{argmin} \ \mathcal{L}(X;\nabla\mathcal{P},\nabla\mathcal{P}^{\prime}) + \alpha TVLoss(X).
\end{eqnarray}
}
The TV-Loss's proportion $\alpha$ is a tunable variable, and momentum can also be incorporated into the optimization process, as detailed in the following sections.

\begin{table}[t]
\caption{Reconstruction Quality of MIP.}
\label{ex1table}
\addtolength{\tabcolsep}{-1pt}
\centering
\scriptsize
\begin{tabular}{cccccccc}
\hline
\multirow{2}{*}{Dataset} & \multirow{2}{*}{Size} & \multicolumn{2}{c}{Soft Prompts}                     & \multicolumn{2}{c}{Double Adapters}                  & \multicolumn{2}{c}{Text Adapter}  \\ \cline{3-8} 
                        &                      & {PSNR $\uparrow$} & {SSIM $\uparrow$} &{PSNR} & {SSIM} & {PSNR} & SSIM  \\ \hline
MNIST                  & 28×28                                      & 15.46                     & 0.723                     & 21.87                     & 0.856                     & 17.38                     & 0.819 \\
CIFAR10                                       & 32×32                                      & 9.18                      & 0.459                     & 11.34                     & 0.572                     & 9.63                      & 0.540 \\
Caltech101                                    & 32×32                                      & 9.01                      & 0.443                     & 10.32                     & 0.481                     & 8.85                      & 0.427 \\
Caltech101                                    & 64×64                                      & 6.77                      & 0.318                     & 10.18                     & 0.464                     & 7.06                      & 0.375 \\ \hline
\end{tabular}
\vspace{-10pt}
\end{table}

In the experimental section on reconstruction quality assessment, for each type of attack module during image restoration, we selected a total of 50 images from 30 different categories across three datasets to average the results. All restored images that converged to a non-random outcome were included in the calculation, which was adopted to mitigate the impact of the inherent randomness associated with our optimization problem on the results. We employed a decaying momentum as a hyperparameter to control TV-Loss during the iteration process, which generally decays from the standard multiplier to 0.001 times. Moreover, class names as the input into the text encoder are obtained directly from the natural semantic names provided by the dataset itself without any additional modifications.

In the gradients obtained through three given attack modes, two are derived from the text modality path, while one originates from a path involving both modalities. According to Proposition~\ref{prop1}, optimizing the gradient from the text path is a first-derivative problem. However, when the loss from adding an image adapter is considered, the problem degrades to a second-derivative issue. At this point, the optimization equation transforms into a problem akin to the DLG standard issue. Therefore, our experiments focused on implementing attacks against text gradients, under which circumstances there is considerable flexibility to replace the text encoder and modify the scale of the soft prompt. Generally, we employed a set of soft prompts with a token length of 512, along with an adapter module composed of two linear layers and two ReLU-activations. When it comes to the implementation part involving the image adapter, given that the internal structure relates to a second-derivative problem, the activation layer of the adapter itself was replaced with a type that is second-differentiable.

\textbf{Reconstruction Quality of MIP.} 
Figure~\ref{ex1} presents the results of reconstruction attacks on soft prompts, with selected images including the ship and automobile categories from CIFAR10, the lamp and dolphin categories from Caltech101, and the number six category from MNIST. Figure~\ref{ex2}, on the other hand, shows the outcomes of reconstructing the gradients originating from the text adapter, selecting the frog and airplane categories from CIFAR10, the ant and wheelchair categories from Caltech101, and the number four category from MNIST. Additionally, we have also implemented the reconstruction based on double adapters, which entails synthesizing the losses from both adapters after the image encoder and the text encoder; these results are displayed in Figure~\ref{ex3}. To recapitulate, Table~\ref{ex1table} shows the aggregated results of the experiments from the three aforementioned sections, with two quality evaluations conducted.

As shown in Table~\ref{ex1table}, we can observe that the reconstruction attack on text gradients can achieve a PSNR above 15db and an SSIM above 0.7 on the MNIST dataset, closely resembling the original images in visual quality and clearly capturing the handwritten digit information. On the CIFAR10 and Caltech101 datasets, our method attains a PSNR above 9db and an SSIM above 0.4 for images of size 32. While these images appear slightly blurred, they predominantly preserve the principal features of the original images in most cases. As the image size increases to 64 or larger, the image quality and convergence probability further decrease due to the multiplying of the search space.

Furthermore, An evident phenomenon is that the images restored using gradients combined from the image path possess higher quality compared to the other two attack modes, which only achieve 80\% of average metrics. However, the second-derivative nature of the former leads to its considerably limited application potential. Meanwhile, employing attack modes targeted at the text encoder can capitalize on their layer-free characteristics, being merely affected by the gradient vanishing issue associated with the image encoder. When utilizing MIP, one can also proactively discard parts of the loss function related to the image adapter to prevent the dispersion of derivatives.

\begin{table}[h]
\vspace{-0.1in}
\caption{Ablation study for MIP.}
\label{ablationtable}
\addtolength{\tabcolsep}{-3pt}
\vspace{0.1in}
\centering
\small
\begin{tabular}{ccccc}
\hline
\multirow{2}{*}{Metric} & {Raw} & {Our} & {+Label }  \\ 
  & Transfer  & Reconst.  & Prediction   \\
\hline
PSNR $\uparrow$                                                     & 0.791\textbackslash{}failed      & 4.152                                    & 8.730                                                                    \\
SSIM $\uparrow$                                                    & 0.042\textbackslash{}failed      & 0.275                                    & 0.582                                                                    \\ \hline
\end{tabular}
\vspace{-0.1in}
\end{table}

\subsection{Ablations}

The three modules of MIP complement each other to produce better results and higher stability. We conducted ablation experiments to determine the indispensable role of each module in image reconstruction. Table~\ref{ablationtable} shows an ablation study for MIP, where each item is a value obtained by averaging the results of 20 images in the experiment. From this table, we can discern that a directly transferred attack strategy is essentially unable to converge in our scenario, primarily due to the influences of feature differences and gradient vanishing. By incorporating measures to circumvent gradient vanishing, especially when reconstructing images from soft prompts, the method achieves a lower recovery quality. Applying the strategy of ground-truth prediction to our scenario can further yield improvements.

\begin{figure}[t]
\centering
\includegraphics[width=0.48\textwidth]{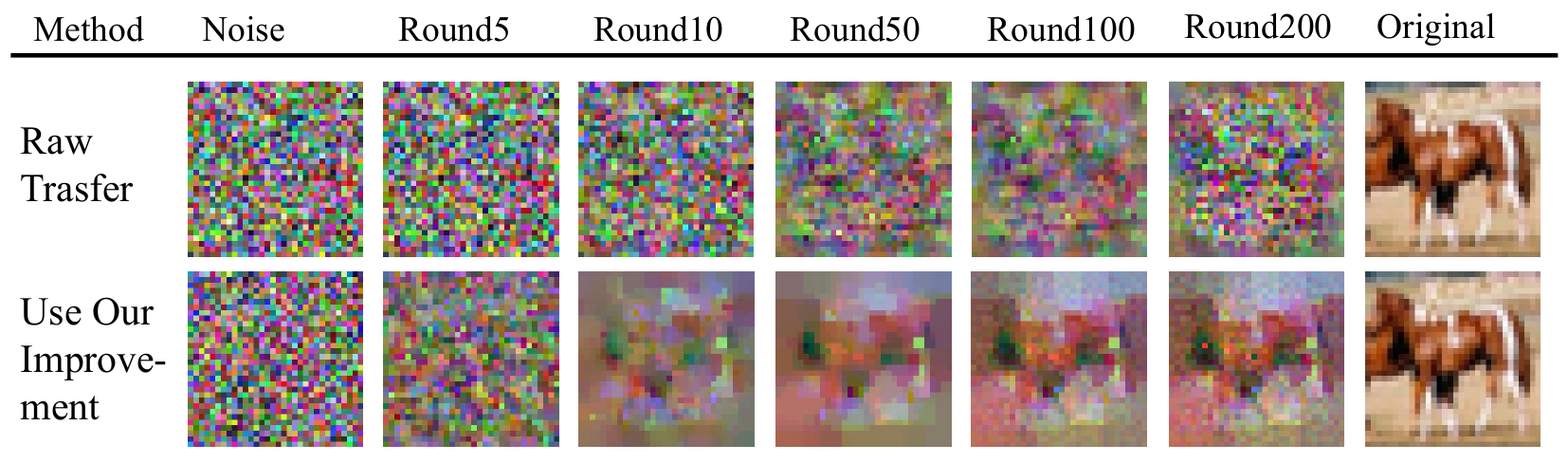}
\vspace{-0.25in}
\caption{Comparison between using our improvement or not.} 
\label{ablation1}
\end{figure}

\textbf{Effectiveness of Gradient Vanishing Avoidance.}
In the previous argument, we have determined the unavoidable rule of the image encoder, so only the structures before the text encoder, such as soft prompts, will be affected by gradient vanishing. As demonstrated in Section~\ref{methodz}, information theft before the text encoder can be reduced to after the text encoder. Without employing this evasion strategy, the success rate of attacks targeting soft prompts leakage is less than 1\%. Figure~\ref{ablation1} illustrates an example for comparison, in which the first line did not employ the improved method we use to circumvent gradient vanishing, whereas the second line represents the scenario where this method is utilized. The results are examples corresponding to the median PSNR in each of the ten experiments. Obviously, it is difficult to reconstruct images without MIP.

\begin{figure}[t]
\centering
\includegraphics[width=0.48\textwidth]{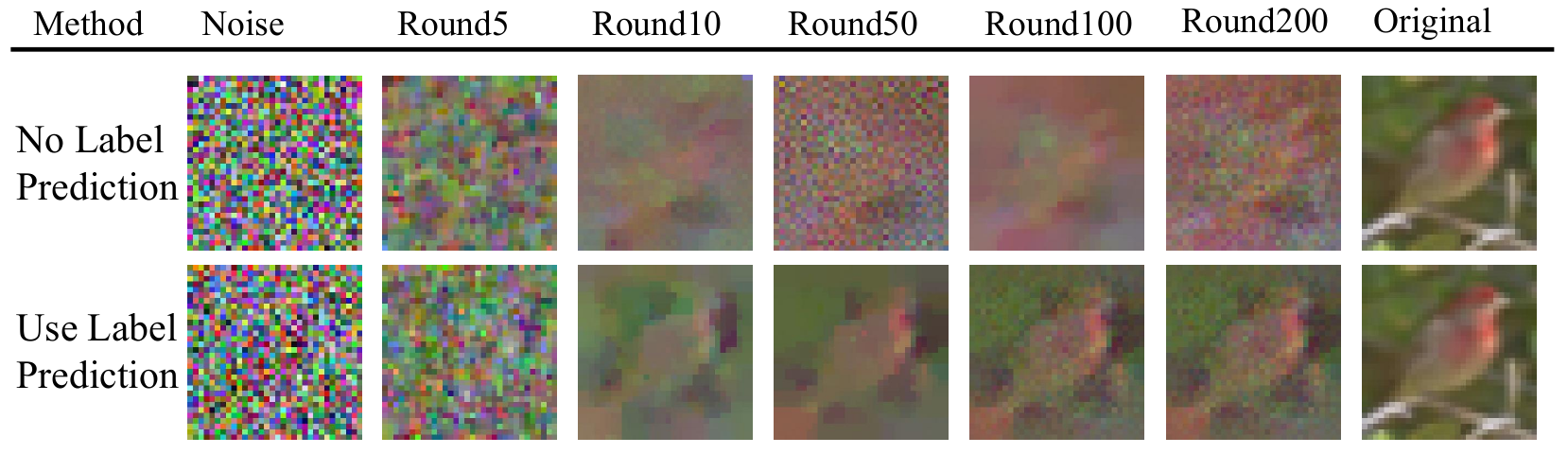}
\vspace{-0.25in}
\caption{Comparison between using label prediction or not.}
\label{ablation2}
\end{figure}

\textbf{Ablation of Label Prediction.}
By introducing label prediction, IDLG significantly improves the convergence quality, which is equally beneficial in our new scenario. Due to the longer derivative chain described in Section~\ref{motiva}, the image reconstruction quality is already lower compared to the original scenario, making it even more necessary to achieve better optimization through ground-truth prediction. Figure~\ref{ablation2} provides an example illustrating the different situations when optimizing Eq.~\ref{eq2} and Eq.~\ref{eq13}. 
In Figure~\ref{ablation2}, two lines represent the median of the PSNR corresponding results of ten optimizations for each of the two modes. Obviously, using label prediction effectively improves stability.

\section{Conclusion}

CLIP performs exceptionally well on multimodal classification tasks, but we have demonstrated that the training approach of fine-tuning PEFT associated with CLIP can be vulnerable to reconstruction attacks in distributed scenarios, exposing the potential privacy leakage issues in such settings. We foremost demonstrated the feasibility of migrating DLG issues to a CLIP-like structure. Then, we proposed MIP that includes label prediction and evasion of long derivative chains in image reconstruction in order to overcome the problems in the new scenario and reconstruct higher-quality images. First, by extending the conclusions of IDLG, MIP constructs an accurate label prediction method from reverse gradient estimation. Second, through experiments on the mitigation of gradient vanishing for the text encoder and the step-like nature of the image encoder, MIP concluded the conditions for reconstruction attacks: the image path is unavoidable, while the structure on the text path has minimal impact on the reconstruction quality. The image quality obtained through these methods far surpasses simple mode migration, and we have also conducted ablation experiments to determine the role of each module.

\bibliography{reference}

\end{document}